%% file: main.tex
\documentclass[runningheads]{llncs}
\usepackage{natbib}
\usepackage{graphicx}
\usepackage{epsfig}
\usepackage{color,soul}
\usepackage{amsmath}
\usepackage{multirow}
\usepackage{multicol}
 \usepackage{float}
\usepackage{pgfplots}
 \usepackage{bigstrut}
 \usepackage{tikz}
\newcommand*\z[1]{$Z_{#1}$}
\DeclareRobustCommand{\r}[1]{#1}

\begin{document}

\title{An Efficient Application of Goal Programming to Tackle Multiobjective Problems With Recurring Fitness Landscapes}
\author{Rodrigo Lankaites Pinheiro\inst{1,2} \and Dario Landa-Silva\inst{2} \and Wasakorn Laesanklang\inst{3} \and Ademir Aparecido Constantino\inst{4}}
\institute{Webroster Ltd., PE1 5NB, Peterborough, UK \\ \email{rodrigo.pinheiro@webroster.com}
\and ASAP Research Group, School of Computer Science, University of Nottingham, UK \\ \email{dario.landasilva@nottingham.ac.uk}
\and Department of Mathematics, Faculty of Science, Mahidol University, Thailand
\\ \email{wasakorn.lae@mahidol.ac.th}
\and Departamento de Inform\'{a}tica, Universidade Estadual de Maring\'{a}, Maring\'{a}, Brazil \\ \email{aaconstantino@uem.br}}
\maketitle

\begin{abstract}
Many real-world applications require decision-makers to assess the quality of solutions while considering multiple conflicting objectives. Obtaining good approximation sets for highly constrained many-objective problems is often a difficult task even for modern multiobjective algorithms. In some cases, multiple instances of the problem scenario present similarities in their fitness landscapes. That is, there are recurring features in the fitness landscapes when searching for solutions to different problem instances. We propose a methodology to exploit this characteristic by solving one instance of a given problem scenario using computationally expensive multiobjective algorithms to obtain a good approximation set and then using Goal Programming with efficient single-objective algorithms to solve other instances of the same problem scenario. We use three goal-based objective functions and show that on benchmark instances of the multiobjective vehicle routing problem with time windows, the methodology is able to produce good results in short computation time. The methodology allows to combine the effectiveness of state-of-the-art multiobjective algorithms with the efficiency of goal programming to find good compromise solutions in problem scenarios where instances have similar fitness landscapes.
\keywords{multi-criteria decision making, goal programming, Pareto optimisation, multiobjective vehice routing}
\end{abstract}

\input{sections/introduction}
\input{sections/vrp}
\input{sections/gp}
\input{sections/proposal}
\input{sections/results}
\input{sections/conclusion}

\setlength{\bibsep}{5pt}
\bibliographystyle{apalike}
{\small
\bibliography{references}
}

\vfill
\end{document}

%% file: sections/introduction.tex
\section{Introduction}

Tackling highly-constrained optimisation problems with many objectives is difficult even with modern multiobjective algorithms \citep{Giagkiozis2012}. In real-world scenarios, decision-makers often benefit from having a set of solutions representing a compromise between the multiple objectives so that they can choose the preferred solution(s). It is often useful to use problem domain knowledge during the optimisation in order to obtain better sets of compromise solutions. For example, in the context of continuous multiobjective optimisation problems, \citet{Fleming2014} estimated Pareto fronts to then obtain values for the decision variables of interesting solutions. Their technique allows to focus the search in sub-regions of the objective space. Another example is the work by \cite{Feliot2016} using a Bayesian model to learn computationally expensive objective functions to then use the estimation model to explore the search space more quickly. 

The multiobjective vehicle routing problem with time windows (MOVRPTW) is a well-know difficult combinatorial optimisation problem that arises in many real-world logistic scenarios \citep{Toth2002}. This problem refers to creating a plan for a fleet of identical vehicles to take goods from a depot and deliver them to customers at various locations. Each customer has certain demand level that needs to be satisfied within a specified time window. Objectives usually considered in the MOVRPTW include among others, the minimisation of number of vehicles and the minimisation of total travel distance by all vehicles.

Due to the high number of constraints and objectives in MOVRPTW scenarios, even state-of-the-art multiobjective algorithms struggle to find good approximations to the Pareto optimal front within reasonable computation time. In logistic scenarios where problems like MOVRPTW arise, it is often the case that problem instances corresponding to a different planning periods share parts of the same data. For example, the same or very similar set of vehicles might be available in each planning period. Also, there might be a set of recurring customer orders that need to be satisfied in the different planning periods. This results in the different problem instances presenting recurring features in their fitness landscapes. Other problems like timetabling and personnel scheduling may also have instances with recurring features resulting in similar fitness landscapes ($\eta$-dimensional surface representing the Pareto front, where $\eta$ is the number of objectives).

Previous work proposed a technique to analyse and visualise complex objective relationships and fitness landscapes in multiobjective problems \citep{Pinheiro2015,Pinheiro2017}. Later, \cite{Pinheiro18} introduced a methodology to exploit the recurring similarity between instances of a multiobjective workforce scheduling and routing optimisation problem, in order to solve instances of the same problem scenario more efficiently. In this methodology, a \textit{pilot} problem instance is solved first using some effective (but not necessarily computationally efficient) multiobjective algorithm to produce an approximation to the Pareto optimal set. Such approximation set is given to the decision-maker so that \textit{target} solutions representing the desired trade-off between the multiple objectives are identified. Then, goal programming is applied with a computationally efficient single-objective solving method, in order to find solutions for other problem instances. In this paper, this methodology is applied to tackle the MOVRPTW in order to further investigate its performance for solving multiobjective problem instances with recurring features. The methodology can be very valuable to facilitate informed decision-making when searching solutions to multiobjective problems. Experiments in this paper are conducted on a set of benchmark instances of the MOVRPTW provided by \cite{Castro-Gutierrez2011}.

Section \ref{s2b} outlines the multiobjective vehicle routing problem with time windows considered here while Section \ref{GP} outlines goal programming. Section \ref{s3} describes the proposed methodology and Section \ref{s4} presents the experimental configuration. Sections \ref{s5b} and \ref{dis} present and discuss the results. Section \ref{s6} concludes the paper and suggests related future research.

%% file: sections/vrp.tex
\section{\r{Multiobjective Vehicle Routing Problem with Time Windows}}\label{s2b}

A Multiobjective Vehicle Routing Problem with Time Windows (MOVRPTW) is defined on a graph $G=(V, E)$ where $V$ is the set of vertices representing the depot (vertex $0$) and the customers (vertices $1 \dots n$) where each customer has a demand $p_i$ ($i=1,\dots,n$). There are $h$ identical vehicles available, each one with capacity $Q$. In this MOVRPTW, $h$ is considered large enough so that as many vehicles as needed are available to create the routing plan. A set of routes served by the set of vehicles should be created in order to satisfy all demands from all customers. All routes must start and end in vertex $0$. The edge set $E$ denotes all possible connections between all vertices. Each edge from vertex $i$ to vertex $j$ has an associated cost, denoted by $c_{ij}$, that represents distance or time for a vehicle to travel between vertices $i$ and $j$. Each customer $i$ must be served during their corresponding time window $[a_{i},b_{i}]$. A waiting time is incurred if a vehicle arrives at time $t < a_{i}$ and hence it must wait until the start of the time window to serve the customer. A delay time is incurred if a vehicle arrives at time $t > a_{i}$ and hence it must start serving the customer immediately. Once the vehicle starts serving the customer, it stays there for $s$ time until the delivery is completed, this is known as the service time.

\citet{Castro-Gutierrez2011} proposed a benchmark set of instances for the MOVRPTW with five minimisation objectives: number of vehicles ($Z_1$), total travel distance by all vehicles ($Z_2$), makespan or travel time of the longest route ($Z_3$), total waiting time for all vehicles ($Z_4$), and total delay time for all vehicles ($Z_5$). They designed their instances based on different characteristics of the problem and each instance is a combination of these features. The features that constitute a problem instance in these benchmarks are:

\begin{itemize}
\item \textbf{Number of customers:} 50, 150 and 250 customers.
\item \textbf{Time window:} five different profiles ($tw0, tw1, tw2, tw3, tw4$) of time windows across a planning period of eight hours. These profiles are defined in terms of minutes from the start of the planning period 0 = 8:00 am, 480 = 4:00 pm, etc.). These five time-window profiles are defined as follows:
\begin{itemize}
    \item{$tw0$:} [0,480], all customers can be served at any time in the day.
    \item{$tw1$:} [0,160],[160,320],[320,480], refers to three types of customers (morning, midday and late).
    \item{$tw2$:} [0,130],[175,305],[350,480], also refers to three types of customers as in profile $tw1$ but with shorter time windows.
    \item{$tw3$:} [0,100],[190,290],[350,480], also refers to three types of customers as in profile $tw1$ but with longer time windows.
    \item{$tw4$:} includes all time-windows from $tw0, tw1, tw2$ and $tw3$, each customer has one of the 10 time window types in the previous profiles.
\end{itemize}
\item \textbf{Demand types:} three types of demand (10, 20, 30) uniformly distributed.
\item \textbf{Vehicle capacity:} the capacity of the vehicles is calculated according to a $\delta$ parameter such that $Q=\underline{D}+\delta/100(\overline{D}-\underline{D})$ where $\underline{D}$ is the maximum single demand among all customers and $\overline{D}$ is the sum of all customer demands. The dataset considers three $\delta$ values ($\delta0=60$, $\delta1=20$, $\delta2=5$).
\item \textbf{Service time:} three values of service time (10, 20, 30) uniformly distributed.
\end{itemize}

For more details of the MOVRPTW described above and a comprehensive study on the multiobjective nature of the problem, please refer to \cite{castro}. There are 45 problem instances and a generator available from https://github.com/psxjpc/MOVRPTW-Generator. The technique to analyse objective relationships described in \citet{Pinheiro2017} was applied to these problem instances and results indicate that indeed they have similar fitness landscapes. This is the case even for instances that have different time window profiles, vehicle capacity and the number of customers. However, in this work, we split the 45 problem instances into three datasets according to the number of customers. This decision was taken because even though the fitness landscapes are similar, the scale of the objective values vary considerably according to the number of customers. Therefore, we have 3 datasets each with 15 problem instances, the set VRP-50 with 50 customers, the set VRP-150 with 150 customers and the set of VRP-250 with 25 customers.

%% file: sections/gp.tex
\section{\r{Goal Programming}}\label{GP}

Without loss of generality, a multiobjective optimisation problem can be written as minimise $F(x) = (f_1(x), f_2(x),...,f_n(x))$ subject to $x \in S$, where $x$ is a solution, $S$ is the set of feasible solutions, $n$ is the number of objectives in the problem, $F(x)$ is the image of $x$ in the $k$-objective space and each $f_i(x)$ is the value of objective $i$ in solution $x$. For two solutions $x$ and $y$, it is said that $x$ dominates $y$, if $ \forall i: f_{i}(x) \leq f_{i}(y)$ and $ \exists j: f_{j}(x) < f_{j}(y)$. Moreover, $x$ is said to be \textit{Pareto Optimal} if it is not dominated by any other feasible solution. Then, the aim is to find the set of \textit{Pareto Optimal} solutions usually called \textit{Pareto Set}. This set contains a number of \textit{non-dominated points in the objective space} creating the \textit{Pareto Front}.

Goal programming is one of the earliest proposed approaches to tackle optimisation problems with multiple objective \citep{CHARNES197739}. Basically, goal programming consists of establishing a specific numeric goal for each of the objectives considered in the problem. Then, search is conducted for a solution in which the weighted sum of deviations in the objective values with respect to the goals is minimised. In order words, goal programming is about establishing a target for each objective and then searching for a solution with objective values as close as possible to those targets. There are three types of goals in goal programming \citep{KORNBLUTH1973193}:

\begin{itemize}
\item Lower bound: defines a lower value for an objective such that solutions that fall below the lower value are penalised.
\item Upper bound: defines an upper value for an objective such that solutions that present higher values than the upper value are penalised.This is the type of goals in the optimisation problem considered here, due to the minimisation nature of all objectives.
\item Strict bound: defines a specific target value such that solutions that present values above or below are penalised. This is applicable when obtaining a solution with a specific target value for a given objective is essential. For example, in the case that solutions using exactly $h$ number of vehicles were required in the MOVRPTW.
\end{itemize}

Once the goals for each objective are set, goal programming techniques derive problem models (LP, MIP, etc.) to find solutions that reach (or are close enough to) the target goals. Several strategies, or goal programming variants, have been presented in the literature. We briefly review the three most widely employed variants \citep{Jones2016}: 

\begin{itemize}
\item \textbf{Weighted GP} \citep{tagkey1991ii}: used when the decision maker is able to assign an \textit{importance} weight to each goal. The objective function for the problem is then a weighted sum of the deviations from the goals.
\item \textbf{Lexicographic GP}: when weighting the goals is difficult, but the decision maker is able to prioritise them, the lexicographic GP technique is commonly applied \citep{Tamiz1995}. The deviations to the target goals are minimised according to defined priority levels such that deviations from a higher level goal are considered infinitely more important that deviations from a lower level goal. 
\item \textbf{Chebyshev GP} \citep{FLAVELL1976731}: consists of minimising the maximum weighted normalised deviation from all the goals, hence promoting solutions that are well-balanced regarding the achievement of the target values.
\end{itemize}

The weighting and lexicographic methods are considered `a priory' approaches in the sense that the decision maker should set a ranking between the objectives before conducting the search for solutions. This is not the case in the Chebyshev method which is an `a posteriori' method because it seeks solutions that are well-balanced in the attainment of all goals so that the decision maker can chose afterwards. In this paper, it is assumed that the decision maker is able to choose a preferred solution from a set of trade-off solutions, instead of being able to establish weights or ranking between the multiple objectives. Hence, only the Chebyshev technique is used later in this work.

A potential issue with goal programming is that it may produce solutions that are not Pareto efficient \citep{Jones2010}. This is especially true when the goals are `pessimistic' and the objectives can be easily achieved. Several methods are proposed to address the issue. Most methods rely on extra information from the decision maker in order to promote the further improvement of certain objectives \citep{Tamiz1999179}. Other methods involve extending the search after the solution is found by the goal programming in order to find dominating solutions \citep{doi:10.1080/03155986.1980.11731801}.

Works in the literature usually describe the application of goal programming using exact methods \citep{Jones2010,Jones2016}. However, many works exist where metaheuristics are employed to solve goal programming models. \cite{doi:10.1080/0305215042000268606} presents a simulated annealing approach to tackle several test problems of preemptive goal programming. \cite{doi:10.1080/13528160500245772} employ a fast converging simulated annealing algorithm to solve a machine-tool selection and operation allocations problem with fuzzy variables. \cite{Ghoseiri20101096} propose a genetic algorithm to tackle a goal programming model for the vehicle routing problem with time windows and \cite{Leung2007} presents a genetic algorithm to tackle a goal programming model for a transportation planning problem with three objectives. Goal programming is a sound approach to tackle the MOVRPTW considered here becasue this technique has been successfully applied to related scheduling and routing problems. For example, it has been applied to nurse scheduling \citep{Azaiez:2005:GPM:1040650.1040656} \citep{doi:10.1080/07408178408974687} and to a version of the vehicle routing problem with soft time-windows \citep{Calvete20071720}.

%% file: sections/proposal.tex
\section{The Efficient GP Methodology}\label{s3}

Figure \ref{fig:estimationoverview} shows the overall concept of the methodology which was originally proposed in \cite{Pinheiro18}. Each of the steps is explained below in reference to the MOVRPTW tackled in this paper. The overall idea is to find a set of compromise solutions for a representative instance of the multiobjective problem in hand. The decision maker then selects from this set a solution that exhibits the desirable qualities in respect of the various objectives, without the need to set weights or priorities for the objectives. The objective values in the selected solutions are set as the targets for goal programming when searching for solutions to the other problem instances (e.g. routing plans for other days in the same problem scenario).

\input{figures/overview}

\begin{enumerate}
\item A \textit{pilot instance} from the dataset with recurring fitness landscape is selected by the decision-maker and solved using multiobjective algorithms to obtain the best possible non-dominated approximation set.
\item The decision-maker chooses a preferred solution $t$ from the obtained non-dominated set. This chosen solution is known as the \textit{target solution} and its objective-vector is denoted by $$\vec{Z^t}=(Z_1^t,Z_2^t,Z_3^t, Z_4^t, Z_5^t)$$
\item Each other instance in the dataset can now be solved with a faster single-objective algorithm using a modified objective function (goal programming variant) aiming to reach the target objective vector.
\item The final solution obtained in Step 3 is presented to the decision maker. The overall advantage of this approach is that Step 1, which is typically computationally expensive, needs to be executed only once for a given representative instance in the problem scenario. Then, other problem instances can be solved faster after the target solution is chosen. 
\end{enumerate}

The modified objective function of Step 3 has an important role in the methodology as it establishes the way in which the search will aim to attain the goals. Three approaches are used here for determining the objective function. The first one is the well known Chebyshev approach. The second one is to derive a weight-vector from the target solution and the approximation set of the pilot instance. The third approach minimises the Euclidean distances to the target objective-vector.   

\subsection{Chebyshev Goal Programming}

Chebyshev goal programming aims to obtain a balanced solution by  minimising the gap to the target of the objective that presents the highest gap, i.e. it seeks to minimise the largest gap to the goals \citep{FLAVELL1976731}. Hence, if the target goals for the objectives are similarly difficult to attain, this technique can obtain a balanced solution. However, if at least one target objective value is more difficult to achieve (i.e. the target goal is too optimistic), the quality of that objective can be a bottleneck for the other objectives because the search will solely focus on improving that objective. We define the Chebyshev objective function for the MOVRPTW as follows:

\input{figures/cheb}

The Chebyshev objective function given by Eq. (\ref{obcv}) is used as the objective function for the MOVRPTW. The main objective is now to minimise $\lambda$, thus finding a well-balanced solution regarding reaching the target values. If all targets are reached, $\lambda$ can assume fractional values and a solution that shows balanced improvements on all objectives may be obtained.

\subsection{Derived Weight Vector}

One problem with the Chebyshev approach is that it does not guarantee Pareto efficiency. However, the optimal solution for a weighted sum objective function (where weights are not simultaneously null) is always Pareto efficient. To derive a weight vector from the target solution, we first convert the approximation set of the pilot instance into a system of linear inequalities. Considering that the approximation set is composed of $k$ objective-vectors ($ \vec{Z}^1$, $ \vec{Z}^2$, $ \dots$, $ \vec{Z}^{5} $), the linear inequalities system can be defined as follows where the aim is to determine the values of $\vec{w}=(w_1,w_2,w_3,w_4,w_5)$:

\begin{figure}[!h]
\begin{flalign} 
\begin{cases}
\vec{w}\vec{Z}^t  \leq  \vec{w}\vec{Z}^1 \\
\vec{w}\vec{Z}^t  \leq  \vec{w}\vec{Z}^2 \\
\qquad \vdots\\
\vec{w}\vec{Z}^t  \leq  \vec{w}\vec{Z}^{k}  \label{sys}
\end{cases}
\end{flalign}
\end{figure}

There is no guarantee that the system of linear inequalities has a solution if the fitness landscape is non-convex, i.e. if no set of weights can be set to achieve some points in the Pareto optimal front. Therefore, instead of finding a solution for the system, we aim to find a weight vector $\vec{w}$ that satisfies the largest number of inequalities. Hence, we define the problem of finding the best weight vector as the following MIP (mixed-integer programming) minimisation problem.

\begin{figure}[!h]
\begin{flalign}
\text{Minimise} & \sum_{j=1}^{k}\overline{x}_j  & \label{wobjective}
\end{flalign}
Subject to
\begin{flalign}
\qquad &\vec{w}\vec{Z^t} - \vec{w}\vec{Z^j} \leq M\overline{x}_j & j = 1,\dots, k& \label{eq:case} \\
&w_i \in (0,1], \overline{x}_j \text{~~~binary} & \begin{cases}
i=1,\dots, 5\\ j=1,\dots,k
\end{cases} & \label{eq:dec} 
\end{flalign}
\end{figure}

The objective function in Eq. (\ref{wobjective}) aims to find a weight vector $\vec{w}$ that minimises the number of linear inequalities in (\ref{sys}) which do not fulfill the condition $\vec{w}\vec{Z^t} \leq \vec{w}\vec{Z^i}$ expressed by constraint (\ref{eq:case}), $M$ is a large constant. Constraint (\ref{eq:dec}) guarantees that zero cannot be chosen as a weight-value (to avoid criteria being removed).

Finally, the weight vector $\vec{w}$ obtained from the MIP model is used in the objective function for the MOVRPTW as given by Eq. (\ref{wobf}).

\begin{figure}[!h]
\begin{flalign}
    \text{Minimise } &
    \qquad \sum_{i=1}^{5} w_i Z_i  & \label{wobf}
\end{flalign}
\end{figure}

\subsection{Euclidean Distances}

We propose an alternative based on the Euclidean distances to the target vector. In essence, this is a method that considers all objectives as equally important. Hence, minimising the Euclidean distances alone does not guarantee Pareto efficiency. In order to mitigate this issue, the proposed method consists of minimising the distances to the target vector for the objectives that are worse than the target. If the current distance for the objectives that are worse than the target vector is small ($<\epsilon$), then the aim is to maximise the distances of the objectives that are better than the target vector.

Henceforth, the objective function in Eq. (\ref{dobjective}) becomes the objective function for the optimisation problem in hand.

\input{figures/distof}

In summary, when the Euclidean distances of the objectives that are worse than the target vector are larger than the given parameter $\epsilon$, the objective function consists of minimising the Euclidean distances ($z$). Otherwise, when $z \leq \epsilon$, the objective consists of maximising the distances for the objectives that are better than the target solution ($z^\prime$). Thus, if the solution has not reached the target, the objective function attempts to close the gap to the target. If the solution is close or better than the target, the objective function attempts to further improve it.

%% file: figures/overview.tex
\begin{figure}[h]
    \centering
    \includegraphics[width=0.60\linewidth]{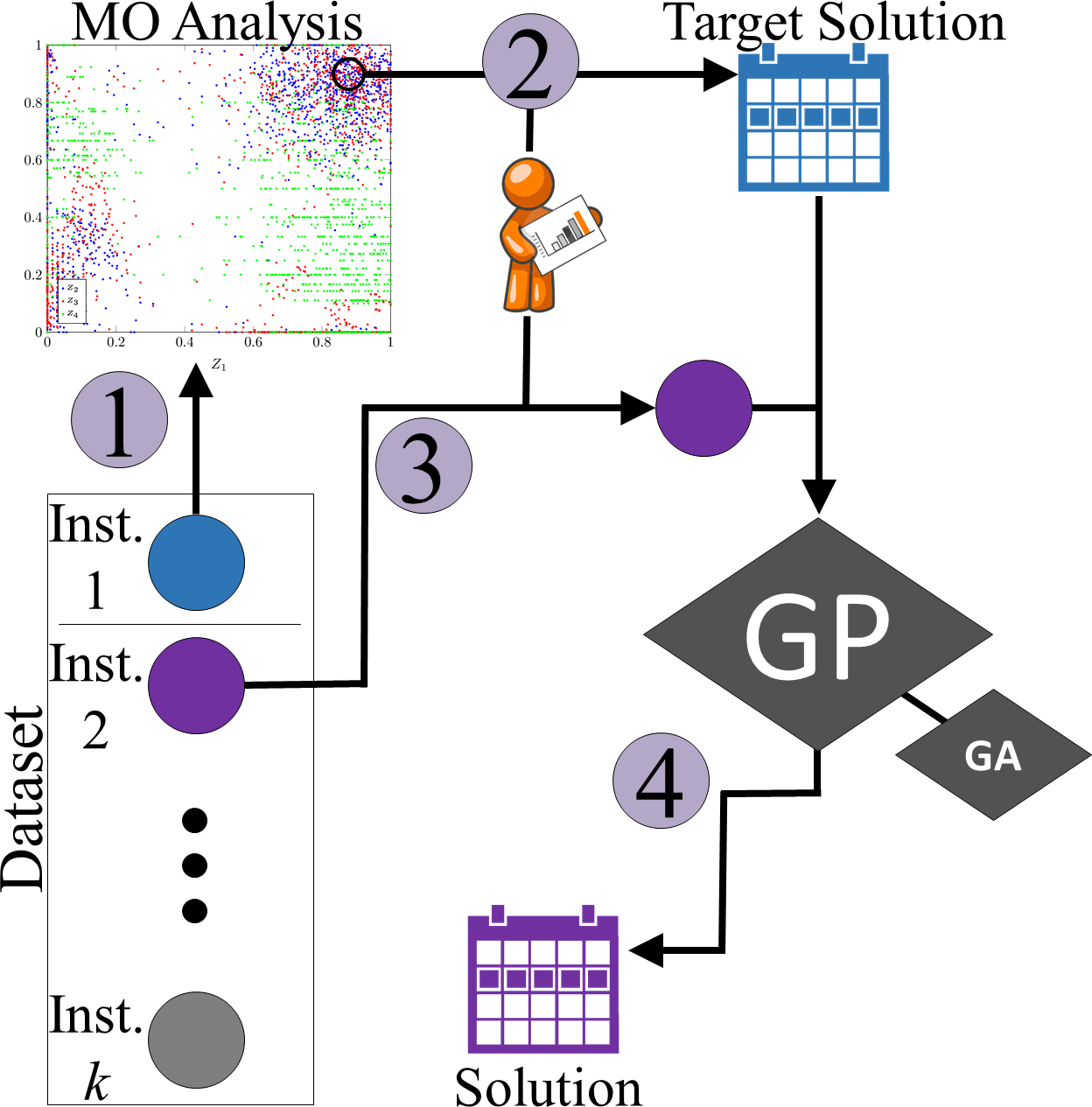}
    \caption{Overview of the methodology as in \cite{Pinheiro18}. The numbered steps are explained in the main text.}
    \label{fig:estimationoverview}
\end{figure}

%% file: figures/cheb.tex
\begin{flalign}
    \text{Minimise~~~~~~~} & \lambda & ~ \label{obcv}\\
    \text{Subject to~~~~~} & & \nonumber\\
        & \frac{Z_1}{Z_1^t} \leq  \lambda \label{cv1}\\
        & \frac{Z_2}{Z_2^t} \leq  \lambda \\
                & \frac{Z_3}{Z_3^t} \leq  \lambda \\
                        & \frac{Z_4}{Z_4^t} \leq  \lambda \\
                                & \frac{Z_5}{Z_5^t} \leq  \lambda \label{cvf} 
\end{flalign}

%% file: figures/distof.tex
\begin{figure}[!h]
\begin{flalign}
\text{Minimize }&\begin{cases}
\displaystyle  z &\qquad\text{if~~~} z >\epsilon  \\
\displaystyle  - z^\prime &\qquad\text{otherwise}
\end{cases} & \label{dobjective}
\end{flalign}
where
\begin{flalign}
\qquad z = &\qquad  \sqrt{ \sum_{i=1}^{5}z_i} &  \\
 z^\prime = &\qquad  \sqrt{\sum_{i=1}^{5}z^\prime_i} & \\
z_i = &\qquad \begin{cases}
(Z_i - Z_i^t)^2 \qquad & \text{if~~~} Z_i > Z_i^t\\
0 &\text{otherwise}
\end{cases} &  \label{culprit}\\
 z^\prime_i = & \qquad \begin{cases}
(Z_i -Z_i^t)^2 \qquad & \text{if~~~} Z_i \leq Z_i^t\\
0 & \text{otherwise}
\end{cases} &
\end{flalign}
\end{figure}

%% file: sections/results.tex
\section{Experimental Configuration}\label{s4}

We applied the proposed methodology to the MOVRPTW datasets. The instances with $\delta0$ and $tw4$ (50-$\delta0$-$tw4$, 150-$\delta0$-$tw4$, 250-$\delta0$-$tw4$ were arbitrarily selected as pilot instances (Step 1 of methodology). Once the Pareto approximation sets were obtained, $k=15$ target vectors were randomly selected (uniformly distributed) from each approximation set and the same target vectors were used for the Derived Weight Vector (WV) objective function, the Euclidean Distances (ED) objective function, and the Chebyshev (CV) objective function.  

Multiobjective algorithms often struggle to find good approximation sets for combinatorial problems with many objectives (more than three) \citep{Giagkiozis2012}. Hence, we resort to a tailored procedure to obtain an improved approximation set. \citet{Fleming2014} state that most multiobjective algorithms can be classified as either Pareto-based or decomposition-based. This study utilises NSGA-II  \citep{Deb2002a} as the Pareto-based algorithm and MOEA/D \citep{Zhang2007} as the decomposition-based one. Thus, \textbf{for each problem instance} the approximation set was obtained (Step 1 of methodology) as described below. The number of solution vectors obtained for each pilot instance was 168 for 50-$\delta0$-$tw4$, 215 for 50-$\delta0$-$tw4$ and 206 for 250-$\delta0$-$tw4$.

\begin{enumerate}
\item run both the NSGA-II and MOEA/D for one million objective evaluations on each possible bi-objective vector (\z{1}, \z{2}), (\z{1}, \z{3}), $\dots$ (\z{4}, \z{5});
\item run both the NSGA-II and MOEA/D for one million objective evaluations on each possible three-objective  vector  (\z{1}, \z{2}, \z{3}), (\z{1}, \z{2}, \z{4}), $\dots$ (\z{3}, \z{4}, \z{5});
\item run both the NSGA-II and MOEA/D for one million objective evaluations on each possible four-objective  vector (\z{1}, \z{2}, \z{3}, \z{4}), (\z{1}, \z{2}, \z{3}, \z{5}), $\dots$ (\z{2}, \z{3}, \z{4}, \z{5});
\item create an archive composed of the non-dominated solutions found in the previous three steps;
\item generate a population of individuals where half of the elements are randomly generated and the other half are randomly drawn from the archive built in the previous step;
\item run both the NSGA-II and MOEA/D four times each, for two million objective evaluations, using the initial population generated in the previous step and the five-objective vector; and
\item compile an approximation set with all non-dominated solutions found in all steps. 
\end{enumerate}

\cite{doi:10.1287/trsc.1030.0057} survey the literature on vehicle routing problem with time windows and show that genetic algorithms are well suited for that problem. Also, our early experiments showed that these algorithms present good enough solutions on these datasets and are simple enough to allow easy replication by other researchers. Hence, for Step 3 of the methodology, the other instances of the MOVRPTW are tackled with a straightforward genetic algorithm (GA) using a direct integer encoding of solutions, uniform crossover, 500 individuals population with a 5\% elite being kept across generations and  a tournament of two individuals employed for the selection mechanism.

\section{Experimental Results}\label{s5b}

\input{figures/accuracyVRP}

First, we show the effectiveness of the derived weight vector obtained from the MIP model in Eqs. (\ref{wobjective})--(\ref{eq:dec}). The effectiveness of a weight vector $\vec{w}$ is given by the percentage of solutions (in the approximation set for the pilot instance) in which $\vec{w}\vec{Z}^t  \leq  \vec{w}\vec{Z}^i$, $i=1,\dots,5$. Hence, if the effectiveness is 100\%, it means that the MIP model found a solution for the inequalities system in Eq. (\ref{sys}).

Figure \ref{fig:accuracyVRP} presents the results of the effectiveness analysis. As it was the case in \cite{Pinheiro18} for another problem, the overall effectiveness of the obtained weight vectors here surpassed 90\%. Pilot instance 50-$\delta0$-$tw4$ presented the best average value of 96\% and 250-$\delta0$-$tw4$ presented the worst with 91.3\%. Hence, in all cases, the MIP model provided good weight vectors to be used by the WV objective function.

Next, we show the results for each group of instances (for 50, 150 and 250 customers) in three charts. The \textit{target achievement} chart displays the percentage of solutions, in the given dataset, that achieved the target value in each objective. The \textit{gap to target} chart contains the average gap to the target solutions for the solutions that did not reach the target. Finally, the \textit{overall comparison} chart displays the average quality of solutions where positive values indicate that, on average, the solutions found are better than the target solution and negative values indicate that the solutions are worse than the target solution.

\input{figures/50-1}

Figures \ref{movns:501}--\ref{movns:503} display the results of applying the implemented GA with all three objective functions (WV, ED and CV) to the other instances of dataset VRP-50. Results comprise the average values of eight runs for each target vector of each problem instance for each objective function.

\input{figures/50-2}

Figure \ref{movns:501} shows that for \z{1} and \z{2} the target achievement is close to 100\% on all three objective functions. On \z{3} the WV objective function noticeably presents the worst results, with only 63\% achievement while the ED and CV objective functions both present similar results with near 80\% achievement. Finally, on \z{4} and \z{5} the ED objective function presents a small advantage and the CV objective function is clearly the worst for \z{5}.

\input{figures/50-3}

Figure \ref{movns:502} reflects the findings of the previous figure where \z{3} shown the lowest overall target achievement. Still, on that objective, the overall gap is below 6\% for the three objective functions, hence when the target was not met, the gap still was small. Noticeably, the ED objective function presents the lowest gaps. Moreover, Figure \ref{movns:503} shows that except for WV on \z{3}, all objective functions on all objectives present improvements over the target solution, noticeably on \z{1}, \z{2} and \z{4} where the solutions found are up to 58\% better than the target.

\input{figures/150-1}

Figures \ref{movns:1501}--\ref{movns:1503} present the results for the larger set VRP-150. On figure \ref{movns:1501}, we see that while on dataset VRP-50 the objective \z{3} presents the worst results, in this dataset the worst results appear on \z{4} with an average of roughly 75\% achievement and, again, the WV objective function presents the worst results. On the other objectives, all objective functions present competitive results.

\input{figures/150-2}

Figure \ref{movns:1502} shows that the gap to the target on solutions that have not met the target is very small -- only on \z{2} the gap is larger than 2\% and only for the CV objective function.

\input{figures/150-3}

Figure \ref{movns:1503} displays the overall quality of solutions. On average, the quality is better on this dataset than on the previous one. With one or more objective functions, on every objective, the overall quality is more than 20\% better than the target. This number increases to nearly 40\% for the WV objective function on \z{1} and \z{2}.

\input{figures/250-1}

Finally, figures \ref{movns:2501}--\ref{movns:2503} present the results for the largest dataset VRP-250. Figure \ref{movns:2501} presents the target achievement. It can be seen that there is a trend, as the size of the datasets increases, the target achievement of \z{1} decreases. In this dataset, the objectives \z{1} and \z{4} presents the worst results. Regarding the objective functions, WV presents the best results for \z{1}. On the remaining objectives, the ED objective function presents the most competitive results.

\input{figures/250-2}

Figure \ref{movns:2502} shows the overall gaps to the target solutions. Clearly, the WV approach gets the worst results, even though the gaps were always below 4.2\%. Noticeably, the CV objective function presents gaps always smaller than 1\%. 

\input{figures/250-3}

Lastly, figure \ref{movns:2503} shows the overall comparison of solutions with their targets. Again, the results show that all objective functions achieved improved results, with the ED edging \z{3}, \z{4} and \z{5} and the WV edging \z{1} and \z{2}.

\section{Discussion}\label{dis}

While the WSRP datasets tackled in \citep{Pinheiro18} arise from real-world scenarios, the MOVRPTW datasets considered here were fabricated for benchmarking purposes. Also, even the largest MOVRPTW scenario is considerably smaller than a medium-sized WSRP. The target achievement for the MOVRPTW here was larger than in the WSRP overall. The best results obtained here were for the smaller MOVRPTW scenarios, while for the WSRP this happened in the larger instances. We speculate that a reason for this is that the largest MOVRPTW datasets are not large enough for the multiobjective algorithms to struggle in finding good approximation sets (as it happened in the larger WSRP datasets). Therefore, as the performance gap between single-objective algorithms and multiobjective algorithms is considerably smaller in the MOVRPTW problem instances, the difficulties of reaching the target vector becomes more evident.

However, the gaps to the targets of objectives that did not meet their targets were considerably lower here than on the WSRP. Also, the CV objective function, while clearly producing the worst results on the WSRP, it achieves competitive results on the MOVRPTW. This could be a consequence of the quality of the target solutions. The multiobjective algorithms were able to obtain approximation sets with fitness landscape closer the fitness landscape of the optimal Pareto front. Also, there is a higher uniformity of the fitness landscapes across instances for these datasets \citep{Pinheiro2017}. All this means that the identified target solutions were realistic, so they could be achieved on every instance. Hence, the CV objective function, which benefits from that, presented good results.

On the MOVRPTW datasets, except for a few exceptions, all objective functions were able to not only reach the target but also to substantially improve all objectives -- also a reflection of the quality of the approximation set obtained by the multiobjective algorithms.

Nonetheless, it is clear that estimating the Pareto front for problem instances that have similar fitness landscape to the pilot instance, is an effective way to tackle the problem. While the multiobjective algorithms required up to four hours to obtain the approximation set for the pilot instance of a dataset, the GA managed to find competitive solutions in minutes. For the majority of the experiments, targets were achieved and the overall quality of results was high.

%% file: figures/accuracyVRP.tex
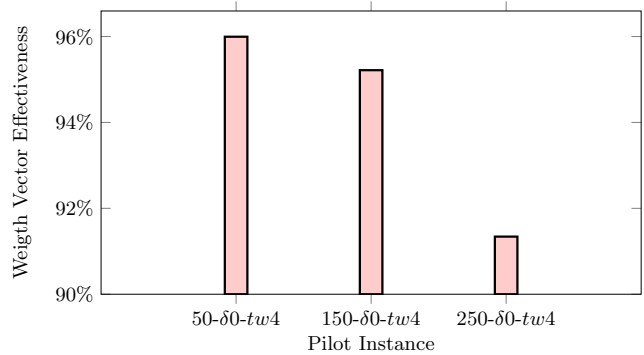
\begin{figure}[!h]
    \centering
{
\resizebox{0.7\linewidth}{!}{%
\begin{tikzpicture}
\begin{axis}[ybar,
width=10cm,
        height=6cm,
legend pos=north west,
visualization depends on=y \as \rawy,
ylabel={Weigth Vector Effectiveness},
tick label style={/pgf/number format/fixed},
yticklabel={\pgfmathparse{\tick*100}\pgfmathprintnumber{\pgfmathresult}\%},
    point meta={y*100},
symbolic x coords={0,50-$\delta0$-$tw4$,150-$\delta0$-$tw4$,250-$\delta0$-$tw4$,1},
xtick=data,
xticklabels = {
            50-$\delta0$-$tw4$,
            150-$\delta0$-$tw4$,
            250-$\delta0$-$tw4$,
        },
        xlabel=Pilot Instance,
xmin=0,
xmax=1,
ymin=0.9,
]

\addplot[draw=black,fill=red!20,nodes near coords, line width=1pt,point meta=explicit symbolic] coordinates{
(50-$\delta0$-$tw4$,0.9599550562)
(150-$\delta0$-$tw4$,0.9521642857)
(250-$\delta0$-$tw4$,0.9134142857)
};
\end{axis}
\end{tikzpicture}
}
}
    \caption{Average percentage of the solutions in the approximation set of each pilot instance in the MOVRPTW datasets such that $\vec{w}\vec{Z}^t  \leq  \vec{w}\vec{Z}^i$.}
    \label{fig:accuracyVRP}
\end{figure}

%% file: figures/50-1.tex
\begin{figure}[!h]
\centering
\pgfplotstableread{
0	1.00000	0.99712	1.00000
1	1.00000	0.98942	0.98942
2	0.63654	0.83077	0.82885
3	0.81154	0.84615	0.82885
4	0.88365	0.91442	0.80769
}\dataset
\center
  \resizebox{0.5\linewidth}{!}{%
\begin{tikzpicture}
\begin{axis}[ybar,
		legend columns=-1, 
		legend style={at={(0.5,1.2)},anchor=north},
        width=6cm,
        height=5cm,
        bar width=0.2cm,
        ymin=0.5,
       	visualization depends on=y \as \rawy,
        x label style={at={(-0.1,-0.05)}},
        xtick=data,
        xticklabels = {
            $Z_1$,
            $Z_2$,
            $Z_3$,
            $Z_4$,
            $Z_5$,
        },
        xticklabel style={yshift=-1.0ex},
        major x tick style = {opacity=0},
        minor x tick num = 1,
        minor tick length=2ex,        
        scaled y ticks = false,
        tick label style={/pgf/number format/fixed},
yticklabel={\pgfmathparse{\tick*100}\pgfmathprintnumber{\pgfmathresult}\%},
        ]
\addplot[draw=black,fill=red!20 ] table[x index=0,y index=1] \dataset; 
\addplot[draw=black,fill=blue!40] table[x index=0,y index=2] \dataset; 
\addplot[draw=black,fill=green!40] table[x index=0,y index=3] \dataset; 
\legend{\tiny{WV},\tiny{ED},\tiny{CV}}
\end{axis}
 \end{tikzpicture}
 }
 \caption{Dataset VRP-50 -- target achievement.}
\label{movns:501}
\end{figure}

%% file: figures/50-2.tex
\begin{figure}[!h]
\centering
\pgfplotstableread{
0	0.00000	0.00001	0.00000
1	0.00000	0.00031	0.00036
2	0.05632	0.02095	0.03442
3	0.00197	0.00000	0.00231
4	0.00414	0.00056	0.01203
}\dataset
\center
  \resizebox{0.5\linewidth}{!}{%
\begin{tikzpicture}
\begin{axis}[ybar,
		legend columns=-1, 
		legend style={at={(0.5,1.2)},anchor=north},
        width=6cm,
        height=5cm,
        bar width=0.2cm,
        ymin=0,     
       	visualization depends on=y \as \rawy,
        x label style={at={(-0.1,-0.05)}},
        xtick=data,
        xticklabels = {
            $Z_1$,
            $Z_2$,
            $Z_3$,
            $Z_4$,
            $Z_5$,
        },
        xticklabel style={yshift=-1.0ex},
        major x tick style = {opacity=0},
        minor x tick num = 1,
        minor tick length=2ex,        
        scaled y ticks = false,
        tick label style={/pgf/number format/fixed},
yticklabel={\pgfmathparse{\tick*100}\pgfmathprintnumber{\pgfmathresult}\%},
        ]
\addplot[draw=black,fill=red!20 ] table[x index=0,y index=1] \dataset; 
\addplot[draw=black,fill=blue!40] table[x index=0,y index=2] \dataset; 
\addplot[draw=black,fill=green!40] table[x index=0,y index=3] \dataset; 
   \legend{\tiny{WV},\tiny{ED},\tiny{CV}}
\end{axis}
 \end{tikzpicture}
 }
 \caption{Dataset VRP-50 -- gap to the target.}
\label{movns:502}
\end{figure}

%% file: figures/50-3.tex
\begin{figure}[!h]
\centering
\pgfplotstableread{
0	0.57650	0.39362	0.32535
1	0.49448	0.30588	0.25621
2	-0.02108	0.08761	0.04947
3	0.33169	0.40823	0.09660
4	0.02435	0.04291	0.00388
}\dataset
\center
  \resizebox{0.5\linewidth}{!}{%
\begin{tikzpicture}
\begin{axis}[ybar,
legend columns=-1, 
		legend style={at={(0.5,1.2)},anchor=north},
        width=6cm,
        height=5cm,
        bar width=0.2cm,
        xmin=-0.5,
        xmax=4.5,
       	visualization depends on=y \as \rawy,
         x label style={at={(-0.1,-0.05)}},
        xtick=data,
        xticklabels = {
            $Z_1$,
            $Z_2$,
            $Z_3$,
            $Z_4$,
            $Z_5$,
        },
        xticklabel style={yshift=-1.0ex},
        major x tick style = {opacity=0},
        minor x tick num = 1,
        minor tick length=2ex,        
        scaled y ticks = false,
        tick label style={/pgf/number format/fixed},
yticklabel={\pgfmathparse{\tick*100}\pgfmathprintnumber{\pgfmathresult}\%},
        ]
\addplot[draw=black,fill=red!20, ] table[x index=0,y index=1] \dataset; 
\addplot[draw=black,fill=blue!40, ] table[x index=0,y index=2] \dataset; 
\addplot[draw=black,fill=green!40, ] table[x index=0,y index=3] \dataset; 
\addplot[draw=black,smooth] 
    coordinates {(-1,0) (6,0)};
    \legend{\tiny{WV},\tiny{ED},\tiny{CV}}
\end{axis}
 \end{tikzpicture}
 }
 \caption{Dataset VRP-50 -- overall comparison.}
\label{movns:503}
\end{figure}

%% file: figures/150-1.tex
\begin{figure}[!h]
\centering
\pgfplotstableread{
0	0.88036	0.89286	0.92143
1	1.00000	1.00000	0.93750
2	0.88304	0.93036	0.83482
3	0.75000	0.78571	0.77054
4	0.98750	0.97500	0.94107
}\dataset
\center
  \resizebox{0.5\linewidth}{!}{%
\begin{tikzpicture}
\begin{axis}[ybar,
		legend columns=-1, 
		legend style={at={(0.5,1.2)},anchor=north},
        width=6cm,
        height=5cm,
        bar width=0.2cm,
        ymin=0.7,
       	visualization depends on=y \as \rawy,
        x label style={at={(-0.1,-0.05)}},
        xtick=data,
        xticklabels = {
            $Z_1$,
            $Z_2$,
            $Z_3$,
            $Z_4$,
            $Z_5$,
        },
        xticklabel style={yshift=-1.0ex},
        major x tick style = {opacity=0},
        minor x tick num = 1,
        minor tick length=2ex,        
        scaled y ticks = false,
        tick label style={/pgf/number format/fixed},
yticklabel={\pgfmathparse{\tick*100}\pgfmathprintnumber{\pgfmathresult}\%},
        ]
\addplot[draw=black,fill=red!20 ] table[x index=0,y index=1] \dataset; 
\addplot[draw=black,fill=blue!40] table[x index=0,y index=2] \dataset; 
\addplot[draw=black,fill=green!40] table[x index=0,y index=3] \dataset; 
\legend{\tiny{WV},\tiny{ED},\tiny{CV}}
\end{axis}
 \end{tikzpicture}
 }
 \caption{Dataset VRP-150 -- target achievement.}
\label{movns:1501}
\end{figure}

%% file: figures/150-2.tex
\begin{figure}[!h]
\centering
\pgfplotstableread{
0	0.01250	0.00798	0.00587
1	0.00000	0.00000	0.00184
2	0.01875	0.01058	0.02528
3	0.00411	0.00000	0.00076
4	0.00850	0.00555	0.02027
}\dataset
\center
  \resizebox{0.5\linewidth}{!}{%
\begin{tikzpicture}
\begin{axis}[ybar,
		legend columns=-1, 
		legend style={at={(0.5,1.2)},anchor=north},
        width=6cm,
        height=5cm,
        bar width=0.2cm,
        ymin=0,     
       	visualization depends on=y \as \rawy,
        x label style={at={(-0.1,-0.05)}},
        xtick=data,
        xticklabels = {
            $Z_1$,
            $Z_2$,
            $Z_3$,
            $Z_4$,
            $Z_5$,
        },
        xticklabel style={yshift=-1.0ex},
        major x tick style = {opacity=0},
        minor x tick num = 1,
        minor tick length=2ex,        
        scaled y ticks = false,
        tick label style={/pgf/number format/fixed},
yticklabel={\pgfmathparse{\tick*100}\pgfmathprintnumber{\pgfmathresult}\%},
        ]
\addplot[draw=black,fill=red!20 ] table[x index=0,y index=1] \dataset; 
\addplot[draw=black,fill=blue!40] table[x index=0,y index=2] \dataset; 
\addplot[draw=black,fill=green!40] table[x index=0,y index=3] \dataset; 
   \legend{\tiny{WV},\tiny{ED},\tiny{CV}}
\end{axis}
 \end{tikzpicture}
 }
 \caption{Dataset VRP-150 -- gap to the target.}
\label{movns:1502}
\end{figure}

%% file: figures/150-3.tex
\begin{figure}[!h]
\centering
\pgfplotstableread{
0	0.37014	0.26069	0.24313
1	0.38086	0.29280	0.22274
2	0.20980	0.24238	0.17840
3	0.17214	0.20336	0.04954
4	0.25293	0.22685	0.13054
}\dataset
\center
  \resizebox{0.5\linewidth}{!}{%
\begin{tikzpicture}
\begin{axis}[ybar,
legend columns=-1, 
		legend style={at={(0.5,1.2)},anchor=north},
        width=6cm,
        height=5cm,
        bar width=0.2cm,
        xmin=-0.5,
        xmax=4.5,
       	visualization depends on=y \as \rawy,
         x label style={at={(-0.1,-0.05)}},
        xtick=data,
        xticklabels = {
            $Z_1$,
            $Z_2$,
            $Z_3$,
            $Z_4$,
            $Z_5$,
        },
        xticklabel style={yshift=-1.0ex},
        major x tick style = {opacity=0},
        minor x tick num = 1,
        minor tick length=2ex,        
        scaled y ticks = false,
        tick label style={/pgf/number format/fixed},
yticklabel={\pgfmathparse{\tick*100}\pgfmathprintnumber{\pgfmathresult}\%},
        ]
\addplot[draw=black,fill=red!20, ] table[x index=0,y index=1] \dataset; 
\addplot[draw=black,fill=blue!40, ] table[x index=0,y index=2] \dataset; 
\addplot[draw=black,fill=green!40, ] table[x index=0,y index=3] \dataset; 
\addplot[draw=black,smooth] 
    coordinates {(-1,0) (6,0)};
    \legend{\tiny{WV},\tiny{ED},\tiny{CV}}
\end{axis}
 \end{tikzpicture}
 }
 \caption{Dataset VRP-150 -- overall comparison.}
\label{movns:1503}
\end{figure}

%% file: figures/250-1.tex
\begin{figure}[!h]
\centering
\pgfplotstableread{
0	0.86964	0.68571	0.77054
1	1.00000	1.00000	0.90000
2	0.86071	0.99911	0.99286
3	0.63125	0.76964	0.76964
4	0.87232	0.99643	0.99196
}\dataset
\center
  \resizebox{0.5\linewidth}{!}{%
\begin{tikzpicture}
\begin{axis}[ybar,
		legend columns=-1, 
		legend style={at={(0.5,1.2)},anchor=north},
        width=6cm,
        height=5cm,
        bar width=0.2cm,
        ymin=0.6,
       	visualization depends on=y \as \rawy,
        x label style={at={(-0.1,-0.05)}},
        xtick=data,
        xticklabels = {
            $Z_1$,
            $Z_2$,
            $Z_3$,
            $Z_4$,
            $Z_5$,
        },
        xticklabel style={yshift=-1.0ex},
        major x tick style = {opacity=0},
        minor x tick num = 1,
        minor tick length=2ex,        
        scaled y ticks = false,
        tick label style={/pgf/number format/fixed},
yticklabel={\pgfmathparse{\tick*100}\pgfmathprintnumber{\pgfmathresult}\%},
        ]
\addplot[draw=black,fill=red!20 ] table[x index=0,y index=1] \dataset; 
\addplot[draw=black,fill=blue!40] table[x index=0,y index=2] \dataset; 
\addplot[draw=black,fill=green!40] table[x index=0,y index=3] \dataset; 
\legend{\tiny{WV},\tiny{ED},\tiny{CV}}
\end{axis}
 \end{tikzpicture}
 }
 \caption{Dataset VRP-250 -- target achievement.}
\label{movns:2501}
\end{figure}

%% file: figures/250-2.tex
\begin{figure}[!h]
\centering
\pgfplotstableread{
0	0.01106	0.01709	0.01100
1	0.00000	0.00000	0.00311
2	0.03768	0.00191	0.00225
3	0.01692	0.00007	0.00005
4	0.04378	0.00076	0.00167
}\dataset
\center
  \resizebox{0.5\linewidth}{!}{%
\begin{tikzpicture}
\begin{axis}[ybar,
		legend columns=-1, 
		legend style={at={(0.5,1.2)},anchor=north},
        width=6cm,
        height=5cm,
        bar width=0.2cm,
        ymin=0,     
       	visualization depends on=y \as \rawy,
        x label style={at={(-0.1,-0.05)}},
        xtick=data,
        xticklabels = {
            $Z_1$,
            $Z_2$,
            $Z_3$,
            $Z_4$,
            $Z_5$,
        },
        xticklabel style={yshift=-1.0ex},
        major x tick style = {opacity=0},
        minor x tick num = 1,
        minor tick length=2ex,        
        scaled y ticks = false,
        tick label style={/pgf/number format/fixed},
yticklabel={\pgfmathparse{\tick*100}\pgfmathprintnumber{\pgfmathresult}\%},
        ]
\addplot[draw=black,fill=red!20 ] table[x index=0,y index=1] \dataset; 
\addplot[draw=black,fill=blue!40] table[x index=0,y index=2] \dataset; 
\addplot[draw=black,fill=green!40] table[x index=0,y index=3] \dataset; 
   \legend{\tiny{WV},\tiny{ED},\tiny{CV}}
\end{axis}
 \end{tikzpicture}
 }
 \caption{Dataset VRP-250 -- gap to the target.}
\label{movns:2502}
\end{figure}

%% file: figures/250-3.tex
\begin{figure}[!h]
\centering
\pgfplotstableread{
0	0.23837	0.19844	0.20253
1	0.36709	0.25866	0.18795
2	0.27977	0.32492	0.28541
3	0.17993	0.26631	0.09464
4	0.18741	0.24293	0.19081
}\dataset
\center
  \resizebox{0.5\linewidth}{!}{%
\begin{tikzpicture}
\begin{axis}[ybar,
legend columns=-1, 
		legend style={at={(0.5,1.2)},anchor=north},
        width=6cm,
        height=5cm,
        bar width=0.2cm,
        xmin=-0.5,
        xmax=4.5,
       	visualization depends on=y \as \rawy,
         x label style={at={(-0.1,-0.05)}},
        xtick=data,
        xticklabels = {
            $Z_1$,
            $Z_2$,
            $Z_3$,
            $Z_4$,
            $Z_5$,
        },
        xticklabel style={yshift=-1.0ex},
        major x tick style = {opacity=0},
        minor x tick num = 1,
        minor tick length=2ex,        
        scaled y ticks = false,
        tick label style={/pgf/number format/fixed},
yticklabel={\pgfmathparse{\tick*100}\pgfmathprintnumber{\pgfmathresult}\%},
        ]
\addplot[draw=black,fill=red!20, ] table[x index=0,y index=1] \dataset; 
\addplot[draw=black,fill=blue!40, ] table[x index=0,y index=2] \dataset; 
\addplot[draw=black,fill=green!40, ] table[x index=0,y index=3] \dataset; 
\addplot[draw=black,smooth] 
    coordinates {(-1,0) (6,0)};
    \legend{\tiny{WV},\tiny{ED},\tiny{CV}}
\end{axis}
 \end{tikzpicture}
 }
 \caption{Dataset VRP-250 -- overall comparison.}
\label{movns:2503}
\end{figure}
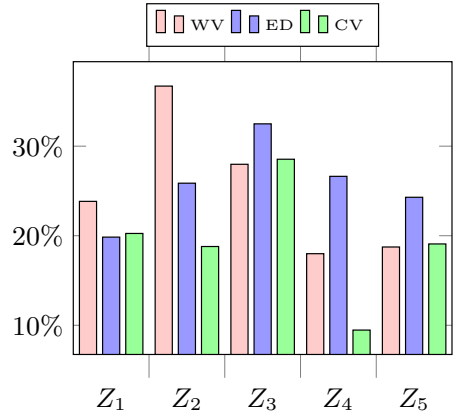

%% file: sections/conclusion.tex
\section{Conclusion}\label{s6}

In this work, we applied a methodology based on goal programming to use efficient single-objective algorithms to solve a multiobjective vehicle routing problem with time windows. The methodology was first presented in \citep{Pinheiro18} and it consists of: 1) solving a pilot instance of the problem using multiobjective algorithms (which are typically computationally expensive) to obtain a good approximation set, 2) having the decision-maker to choose preferred target compromise solutions, and then 3) employing goal programming to solve other instances of the same dataset using the selected solutions in 2) as the target. Three different objective functions were used to guide the search for the target solutions with goal programming. One is the Chebychev approach that seeks to achieve a solution balanced on all the objective targets. Another one is minimising a weighted function derived from the target solution. The third approach is to use the Euclidean distance to drive the search guided by the target solution.

This methodology was first applied by \cite{Pinheiro18} to real-world instances of a Workforce Scheduling and Routing Problem (WSRP) in the home healthcare sector. In the present paper, the methodology has been further tested by applying it to a different multiobjective problem arising in logistic operational scenarios, the Multiobjective Vehicle Routring Problem with Time Windows (MOVRPTW). In both of these problem scenarios, instances usally arise from different planning periods and hence they present similarities in the fitness landscapes. This is because usually in this type of real-world problems, instances have the same partial data (e.g. same fleet of vehicles or same set of workers). This paper has shown that the proposed technique is an effective and efficient approach to tackle real-world multiobjective highly-constrained combinatorial optimisation problems, by combining the effectiveness (but often computationally expensive) of state-of-the-art multiobjective algorithms with the efficiency of well-targeted single-objective optimisation through goal programming. For this, the multiobjective analysis technique proposed by \citep{Pinheiro2015, Pinheiro2017} offers an effective tool to analyse the relationships between objectives in multiobjective optimisation problems and determine the degree of similarity in the fitness landscape of different problem instances.

For future research, it would be interesting to investigate if other approaches besides the Chebychev, derived weighted function and Euclidean distance approaches, would be more effective across different multiobjective problems. Perhaps an even more interesting but also more challenging future research would be to develop adaptive objective functions that change the search direction as the search progresses and in reaction to the fitness landscape features.